\documentclass{article}

\PassOptionsToPackage{numbers, compress}{natbib}
\usepackage[dandb]{neurips_2025}

\usepackage[utf8]{inputenc} 
\usepackage[T1]{fontenc}    
\usepackage{hyperref}       
\usepackage{url}            
\usepackage{booktabs}       
\usepackage{amsfonts}       
\usepackage{nicefrac}       
\usepackage{microtype}      
\usepackage{xcolor}         
\usepackage{graphicx}
\usepackage{natbib}

\usepackage{amsmath} 
\usepackage{longtable}
\usepackage{multirow}
\usepackage{listings}
\usepackage{wrapfig}
\usepackage{tabularx}
\usepackage{tcolorbox}
\usepackage{lipsum}
\usepackage{float}
\usepackage{changepage}
\usepackage{placeins}
\usepackage{svg}
\usepackage{tikz}
\usepackage{enumitem}
\usepackage{subcaption}
\usepackage{caption}
\usepackage{pifont, ragged2e}
\usepackage{rotating}

\title{SridBench: Benchmark of Scientific Research Illustration Drawing of Image Generation Model}

\author{%
  Yifan Chang$^{1,2}$\thanks{Equal contributions.  $\dagger$Corresponding author}\quad 
  Yukang Feng $^{2,3}$\footnotemark[1]  \quad 
  Jianwen Sun$^{2,3}$\footnotemark[1] \quad 
  Jiaxin Ai$^{2,4}$\quad \and
  \textbf{Chuanhao Li}$^{5}$ \quad
  \textbf{S. Kevin Zhou}$^{1}$ \quad
  \textbf{Kaipeng Zhang}$^{2,5\dagger}$ \quad
  \\[2mm]
    $^1$ University of Science and Technology of China \quad
    $^2$ Shanghai Innovation Institute \\
    $^3$ Nankai University \quad
    $^4$ Wuhan University \quad
    $^5$ Shanghai AI Laboratory \\
    \vspace{-5mm}
}

\begin{document}

\maketitle

\begin{abstract}
Recent years have witnessed rapid progress in AI-driven image generation. Early diffusion-based methods focused on perceptual quality, while more recent multimodal models—such as GPT-4o-image—have begun integrating high-level reasoning into the generation process, demonstrating stronger capabilities in semantic understanding and structural composition.
Scientific research illustration generation stands at the forefront of this evolution. Unlike general-purpose image synthesis, this task requires models to accurately interpret complex technical descriptions and transform abstract structures into clear and standardized visual representations. It is significantly more knowledge-intensive than ordinary image generation. According to recent surveys, producing a single research figure typically demands several hours of manual work, often accompanied by expensive software tools and repeated revisions. Automating this process in a controllable and intelligent manner would therefore yield substantial practical benefits. However, no benchmark currently exists to systematically evaluate AI performance on this task.
To address this gap, we present SridBench, the first benchmark designed to assess multimodal models on scientific figure generation. It consists of 1,120 instances via human experts and multimodal large language models(MLLM) on the authoritative scientific paper website which spanning 13 disciplines under natural science and computer science, with each sample evaluated along six well-designed dimensions including semantic fidelity and structural accuracy. 
Our experiments show that even state-of-the-art models like GPT-4o-image fall far short of human-level performance.  At present, the lack of text and visual information and scientific errors are the main bottlenecks of GPT-4o-image for scientific illustration drawing, underscoring the need for further advances in reasoning-driven visual generation.
\end{abstract}

\section{Introduction}

Scientific illustration are essential tools for communicating research findings. They translate complex frameworks, data, and experimental procedures into intuitive visuals, playing a central role in both scholarly publications and scientific discourse. However, creating high-quality illustrations is time-consuming, labor-intensive, and often requires proficiency in both domain-specific knowledge and design tools. This bottleneck limits productivity and slows down the rapid iteration cycles demanded by modern research workflows.

Recent advances in generative AI have made automatic image generation a promising direction in graphic design. Diffusion models ~\citep{DSM, ddpm, diffusion}(e.g., Stable Diffusion~\citep{sd3}, DALL·E~\citep{dalle2, dalle3}, Flux~\citep{flux}) have demonstrated impressive capabilities in visual fidelity and stylistic diversity. Autoregressive vision-language models (e.g., Emu3~\citep{emu3}, NAR~\citep{nar}, VAR~\citep{VAR}, Janus-Pro~\citep{januspro}) further extend the boundaries by improving semantic alignment between textual inputs and visual outputs, especially in open-ended generation tasks. However, their limited reasoning ability in complex application scenarios remains a major constraint on generative performance. With the advancement of reasoning capabilities in language models, models like GPT-4o-image~\citep{openai2025gpt4o}, which incorporate chain-of-thought reasoning and stronger multimodal foundations, mark a shift towards more controllable and content-aware generation. In principle, such models can be used to generate scientific illustrations directly from textual descriptions.

\begin{figure}[htbp]  \centering\includegraphics[width=1\textwidth]{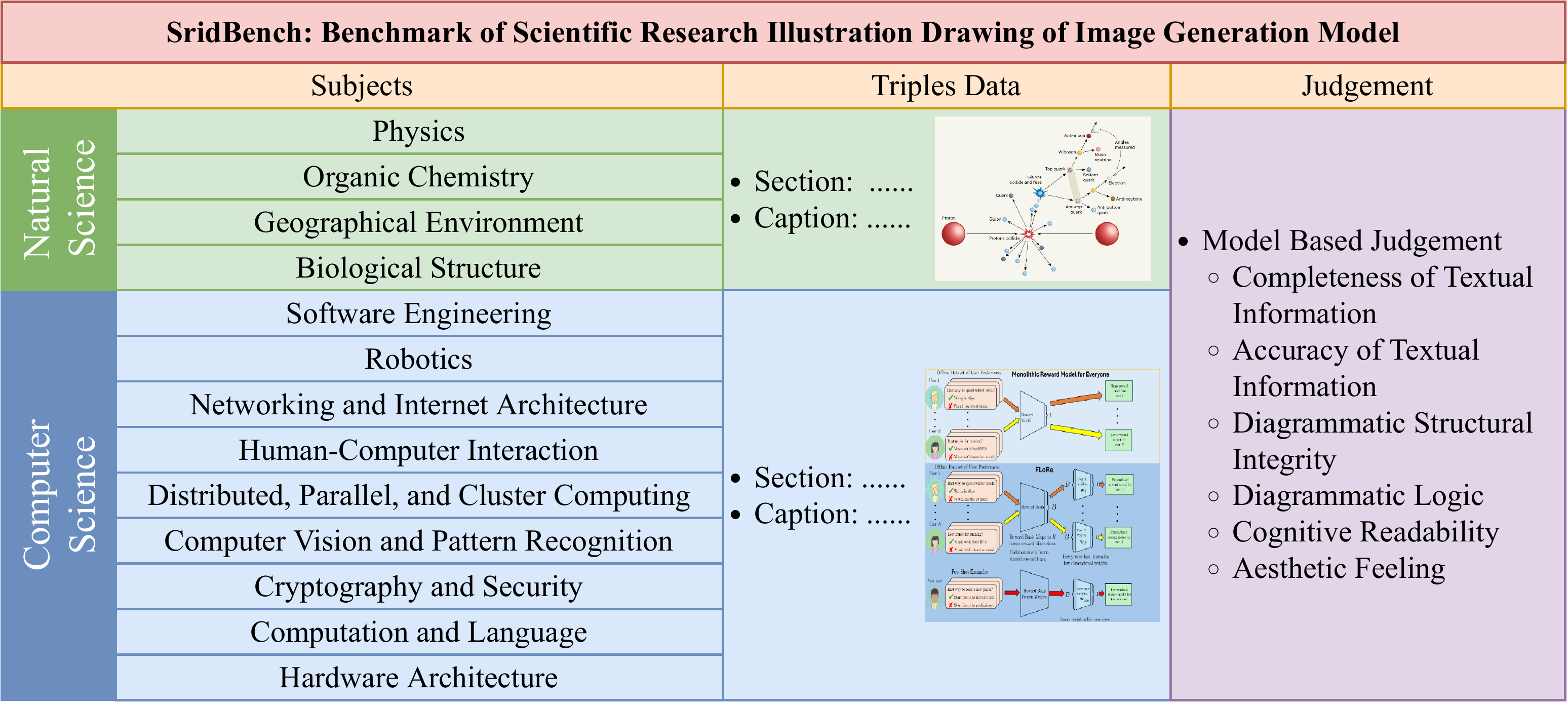} \caption{ General description of SridBench.   We collected triple data from 13 directions in natural science and computer science, and designed 6 evaluation metrics } \label{first}   \end{figure}

Currently, research on AI-assisted scientific illustration remains in its early stages and is mainly focused on benchmarking the understanding capabilities of multimodal models (e.g., SciFIBench~\citep{SciFIBench}, ScImage~\citep{ScImage}). There is a noticeable lack of evaluation frameworks for assessing the ability of models to generate scientific diagrams. A key open question is how to objectively and systematically evaluate the quality of scientific illustrations produced by generative models. To fill this gap, we introduce SridBench, the first benchmark specifically designed to evaluate the capability of multimodal models to generate scientific graphics from textual descriptions. This dataset includes 1,120 generation instances collected from peer-reviewed publications across 13 academic disciplines as shown in Fig.\ref{first}. For systematic evaluation, each instance is annotated and assessed along six dimensions, supporting both human and automated evaluation protocols.

Additionally, we conduct extensive benchmarking of a wide range of generative models. Results reveal a significant performance gap between current models and expert-created graphics. Even the best-performing model in our study, GPT-4o-image, achieves an average score of fair level. Semantic understanding emerges as the primary bottleneck. Open-source models perform worse, with average scores close to 1, and proprietary models like Gemini-2.0-Flash only reach a score of 1.0, highlighting the considerable room for improvement in this domain.

In summary, this work makes the following key contributions:
\begin{enumerate}
    \item SridBench, a new benchmark dataset featuring 1,120 high-quality generation instances from real-world scientific literature and spanning 13 disciplines under natural science and computer science;
    \item A multi-dimensional evaluation protocol assessing figure quality along six dimensions, supporting both human and automated scoring;
    \item A comprehensive empirical study providing the first systematic comparison of representative generative models in the context of scientific illustration, revealing actionable research challenges.
\end{enumerate}

\section{Related work}

In the task of image generation, the current mainstream AI generation models mainly include two categories: diffusion models and autoregressive models. They have continuously made breakthroughs in generation quality, control ability, and multimodal understanding, providing an important foundation for application scenarios such as scientific research illustrations.

\textbf{Diffusion models: }By simulating the process of gradually adding noise to data and then reverse-restoring it, diffusion models~\citep{sd3} have made significant progress in generation accuracy and controllability in recent years. 
Represented by the Stable Diffusion series, its latest version, Stable Diffusion XL~\citep{sdxl}, has been included in the MLPerf~\citep{MLPerf} Inference v4.0 benchmark test, demonstrating its powerful performance in high-quality image generation. 
Stable Diffusion series have demonstrated its powerful performance in high-quality image generation. 
DALL·E3~\citep{dalle3} performs well in design-related tasks such as DEsignBench~\citep{DEsignBench}, indicating its strong text-to-image alignment ability. The FLUX series of models strikes a balance among image resolution, generation speed, and cost-efficiency, being particularly suitable for high-resolution image generation tasks. Although diffusion models have achieved remarkable results in visual generation, their performance in complex scenarios still faces challenges. Especially in scientific research drawing tasks that require strict semantic control and structural constraints, there is still room for improvement in their context understanding ability and structural controllability.

\textbf{Autoregressive models:} In contrast, autoregressive models~\citep{vilau, emu2, emu3} predict the pixels or feature positions in an image step by step, enabling more accurate alignment with the semantics of the input text while ensuring consistency. Emu3 has achieved leading results in text-to-image tasks such as T2I-CompBench~\citep{t2icompbench}, showing its high-fit ability in multimodal understanding and image generation tasks. Janus-Pro performs superiorly in the multimodal consistency of text-to-image, especially demonstrating the unique advantages of the autoregressive structure in detail restoration and instruction response. Currently, autoregressive models are widely regarded as a generation paradigm more suitable for handling high-semantic-density inputs and have important potential in scientific research-related image generation tasks.

\textbf{Reasoning ability:} Recent research trends~\citep{Qwen2-VL, Qwen2.5-VL, internvl, internvl2.5} indicate that reasoning ability is a key factor affecting whether a generation model can be adapted to complex scientific research scenarios. Starting from the o1 model released by OpenAI, large models that introduce the ``Chain-of-Thought (CoT)''~\citep{cot} mechanism (such as DeepSeek-R1~\citep{deepseekrl}, QwQ~\citep{qwen2024qwq}, Doubao, etc.) have demonstrated powerful capabilities in multimodal reasoning tasks. These models can not only analyze the deep-level logical relationships in the input text but also generate responses that conform to the context semantics in complex scenarios.
Furthermore, new-generation models such as GPT-4o-image have deeply integrated advanced reasoning ability with image generation ability for the first time, possessing the ability to "understand scientific research texts" and generate scientific research illustrations with reasonable structures and accurate content accordingly. The evolution of this paradigm indicates that generation models are moving from "art-level" to "scientific-research-level", providing a new path to solve the understanding and control problems in scientific research illustration generation.

\textbf{Research Gaps:} Although the development of generative models and reasoning models has laid a good foundation for scientific research diagram tasks, there is currently a lack of systematic evaluation framework to measure their actual performance in this specific scenario. Relying on human evaluation lacks a unified standard, making it difficult to objectively quantify the performance of models in terms of semantic accuracy, structural rationality, and aesthetic quality. At the same time, most of the current research efforts are focused on the understanding of scientific images and the generation of image captions (such as SciFIBench~\citep{SciFIBench}, FigCaps-HF~\citep{FigCapsHF}, etc.~\citep{ChartBench}). The field of evaluating the generation of scientific research drawings is almost blank. Therefore, this paper focuses on constructing a multi-dimensional scientific research drawing evaluation system and systematically comparing the performance of different types of generative models, providing theoretical support and practical basis for the implementation of multimodal generation technology in the scientific research visualization scenario.

\section{Method}
\begin{figure}[htbp]  \centering\includegraphics[width=1\textwidth]{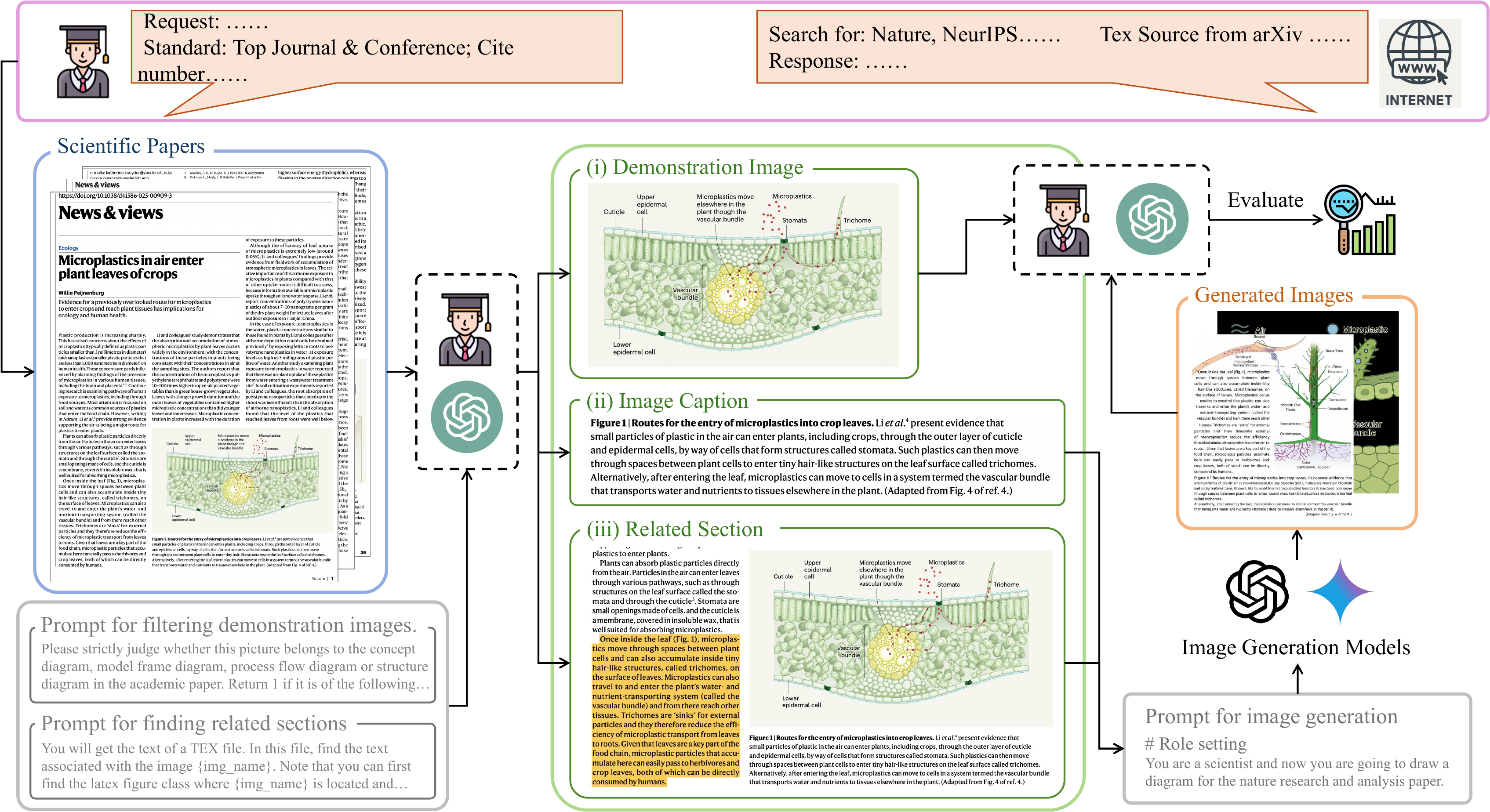} \caption{The framework of our Benchmark of Scientific Research
Illustration Drawing of Image Generation Model. As can be seen from the framework, human experts set the standards for batch downloading and filtering paper data from the Internet. MLLM and human experts work together to screen triplet data to ensure the authority and scientific nature of the data. At the same time, we use the MLLM which is consistent with the human preference and evaluation for automatic scoring.
} \label{framework}   \end{figure}

In order to test the scientific research drawing ability of image generation models, we collect and carefully select the scientific research illustration drawing data, and set the process and standard for scientific research drawing evaluation. This process can be seen in Fig.\ref{framework}. We collect data in two disciplines: \textbf{Computer Science} and \textbf{Natural Science}. Prompts mentioned in this section are shown in Appendix.\ref{prompt}.

\subsection{Collection and structuring of data}

 We collect papers and filter data on professional and authoritative paper websites in these two disciplines. The filtering process is to use Multimodal Large Language Models(MLLMs) to determine whether the illustration in the paper is a schematic diagram or illustration (rather than a real photo, experimental results diagram, and statistical data analysis diagram) , find and extract the caption and section. In this way, we can get a lot of structured triple data: {image, caption, section} . Human experts will sift through the resulting triplet data. Specifically, the goal illustration in the triple should be clear, scientific, rigorous, and has a certain degree of expressiveness. The text should also be able to support and cover the elements that generate the illustration.

 We try to find papers in the top journals and conferences in all directions that have recently been published. ArXiv can help us get some of the computer science papers TeX source files. This is necessary for us to construct triples because we can obtain the LaTeX expression of the formula in the original paper. ArXiv's API makes it easy to get the TEX files for a large number of papers in a given direction. However, not all papers from top conferences and journals are submitted to arXiv. Therefore, we screened arXiv preprint papers that were not published in formal journals and conferences. For articles with fewer than 25 citations, we remove them directly. We then invited human experts to assess the content, quality, and quality of the illustrations. Only papers and illustrations considered to be of high quality and scientific quality by human experts will be used to construct triples. For natural science papers, we crawl from the Nature website, which ensures the quality and authority of the data.

\subsection{ Automatic generation and scoring }

 Once we have the triples, we fill the text of the caption and section into a well-designed prompt template, and then use the image generation model to draw the research illustration. Using MLLM's API, we implement a batch and automatic image generation process.

After obtaining the generated illustrations, we compare them to the images in the triple and score them using MLLM. We designed 6 scoring dimensions for scientific illustration. First, in order to measure the scientificalness and completeness of visual elements (such as organelles, molecular structure, neural network modules) , we set up two indicators: ``Diagrammatic structural integrity''and ``Diagrammatic logic''. Secondly, in order to measure the quality of the text in the illustrations, we set two indicators of ``Completeness of textual information'' and ``Accuracy of textual information''. Finally, we design ``Cognitive readability'' and ``Aesthetic feeling'' to evaluate the quality of the generated results as a whole. These six metrics are written into the prompt using MLLM scoring on a scale of 1 to 5 (1: fail, 2: poor, 3: fair, 4: good, 5: excellent)

\section{Experiments}

\subsection{Experimental Setup}

\paragraph{Model}
Due to the limitation of input length, only three image generation models are chosen by us to evaluate the capability of scientific research drawing, which are GPT-4o-image, Gemini-2.0-Flash~\cite{gemini-2-0-flash} and Emu-3.  We quantitatively analyzed GPT-4o-image and Gemini-2.0-Flash because Emu-3 takes too long time to generate all images. GPT-4o is used to judge the generation results of those models.

\paragraph{Data}
Computer Science data are collected on arXiv and in top journals and conferences of computer science, while 
Natural Science data are collected on Reviews $\&$ Analysis section on the website of Nature (https://www.nature.com/nature/reviews-and-analysis). Since there is no section in the above paper, we choose the paragraph where the illustration is located as the section. 

Inspired by the classification on arXiv, we set nine specific directions in the collection of Computer Science data: Software Engineering, Robotics, Networking and Internet Architecture, Human-Computer Interaction, Distributed, Parallel, and Cluster Computing, Computer Vision and Pattern Recognition, Cryptography and Security, Computation and Language, and Hardware Architecture. 100 triples are selected in each direction. Meanwhile, we carefully select 220 triples from the latest reviews and analysis section of Nature. The selection process is done by experts in various fields. This careful selection process ensures the quality, timeliness, and diversity of the data.

In addition, we use GPT-4o to further categorize images by scene and function to facilitate the analysis of more results. Eight categories are chosen to categorize computer science images: software design, noun classification, mathematical structure, hardware design, engineering system design, algorithm flow, AI model, and other types. Natural science images are categorized into four types: Physics diagram, Organic Chemistry diagram, Geographical Environment diagram, and Biological Structure diagram. 


    
\subsection{Overall Evaluation}

\begin{figure}[htbp]  \centering\includegraphics[width=1\textwidth]{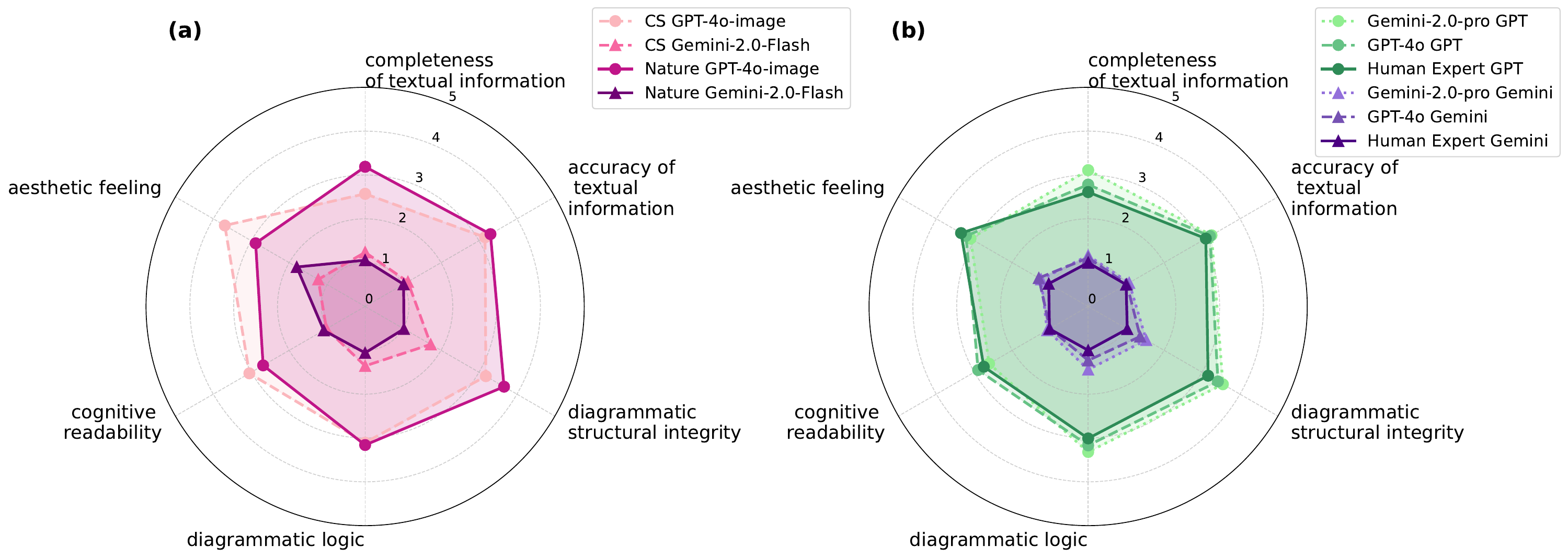} \caption{(a). On the computer science and natural science data, the average score of GPT-4o-image and Gemini-2.0-Flash scores in the six major indicators judged by GPT-4o. (b). For images generated by GPT-4o-image and Gemini-2.0-Flash, the comparison of score judged by  Gemini-2.0-pro, GPT-4o and human expert.} \label{radar}   \end{figure}

As we can see in Fig.\ref{radar}(a), Gemini-2.0-Flash scored less than 2 on each of these measures, meaning that the model had little or no ability to draw professional illustrations for scientific research papers. In the category ``diagrammatic structural integrity'', Gemini-2.0-Flash earned the highest score of all. This shows that Gemini-2.0-Flash has a basic understanding of the basic style and frame structure of scientific drawing. However, in terms of concrete content, as well as scientific logic and deduction, Gemini-2.0-Flash has no such ability at all.

 GPT-4o-image fully demonstrates its superiority over Gemini-2.0-Flash in this task. Every subject, computer science or natural science, is rated at around 3, with most scoring above that mark. This means that GPT-4o-image's scientific mapping capabilities are at a level of basic eligibility that humans would consider acceptable.

  At the same time, we selected 50 natural science and 50 computer science triplets from the dataset, allowing Gemini-2.0-pro, GPT-4o and human experts to independently rate them simultaneously. The results, shown in Fig.\ref{radar}(b), show that the GPT-4o scores are broadly in line with those of human experts, while  Gemini-2.0-pro scores are significantly biased relative to human scores. Therefore, we use GPT-4o for automated scoring. However, GPT-4o is still slightly overrated in terms of completeness and accuracy compared to human expert ratings.

\subsection{Evaluation on Natural Science Data}

 \begin{figure}[htbp]  \centering\includegraphics[width=1\textwidth]{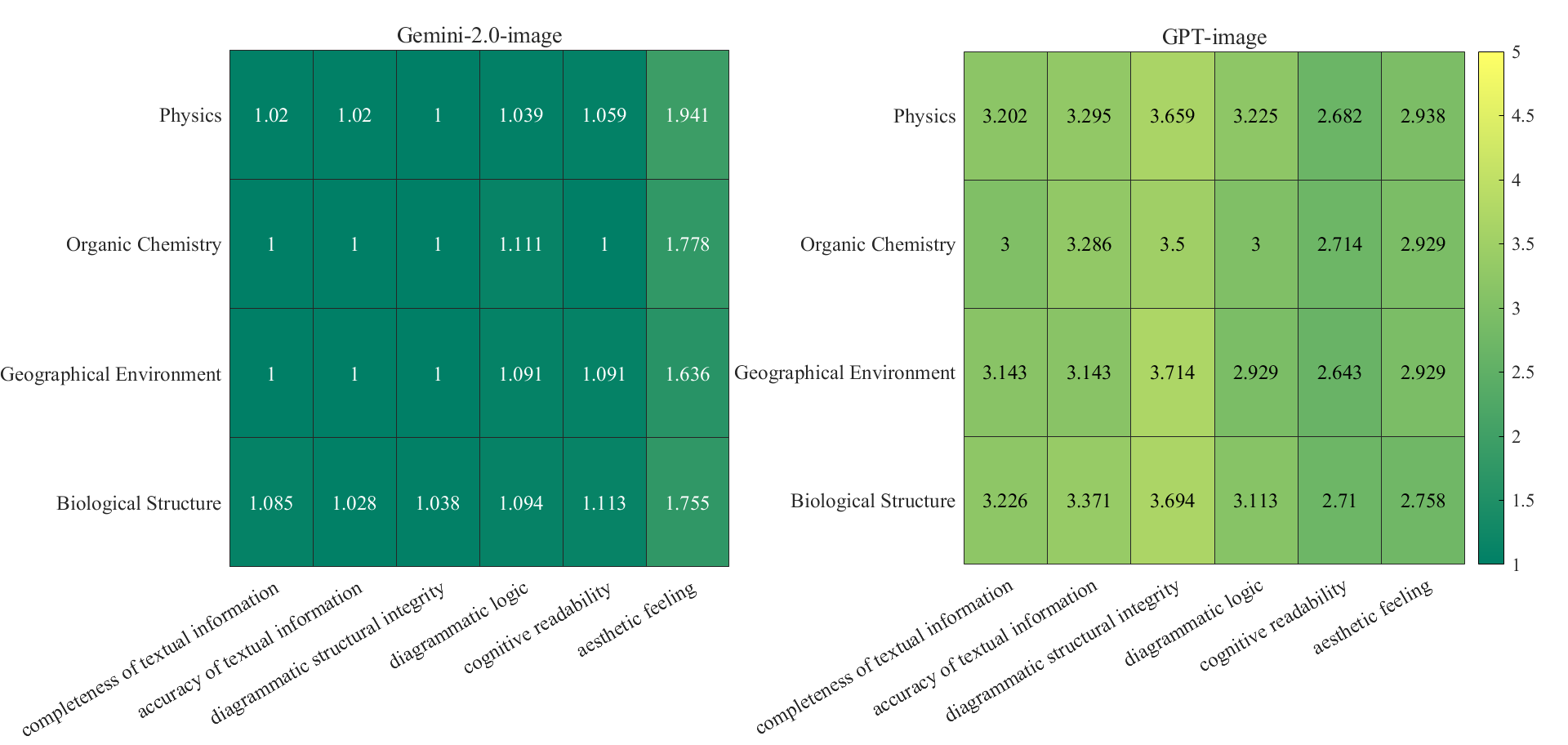} \caption{On different subjects of natural science data, the average score of GPT-4o-image and Gemini-2.0-Flash scores in the six major indicators.
} \label{ns_hm}   \end{figure}

The performance of the two models on natural science data is demonstrated in Fig.\ref{ns_hm}. More specifically, the integrity of GPT-4o-image generated image elements (e.g., cell structure, sensing instrument structure) is significantly higher than the integrity of the text elements (whether or not it completely covers all the information in the reference image). GPT-4o-image can guarantee the accuracy of expressing text, although it cannot express text information completely. As can be seen, the accuracy of the text information on this side of the score is higher than the integrity. However, in terms of logic, simplicity and aesthetics, GPT-4o-image scored below average. This means that there is still much room for improvement in the overall look and feel of natural science image rendering. Overall,  GPT-4o-image does not show a significant gap in competence between different disciplines of the natural sciences .

\subsection{Evaluation on Computer Science Data}

\begin{figure}[htbp]  \centering\includegraphics[width=1\textwidth]{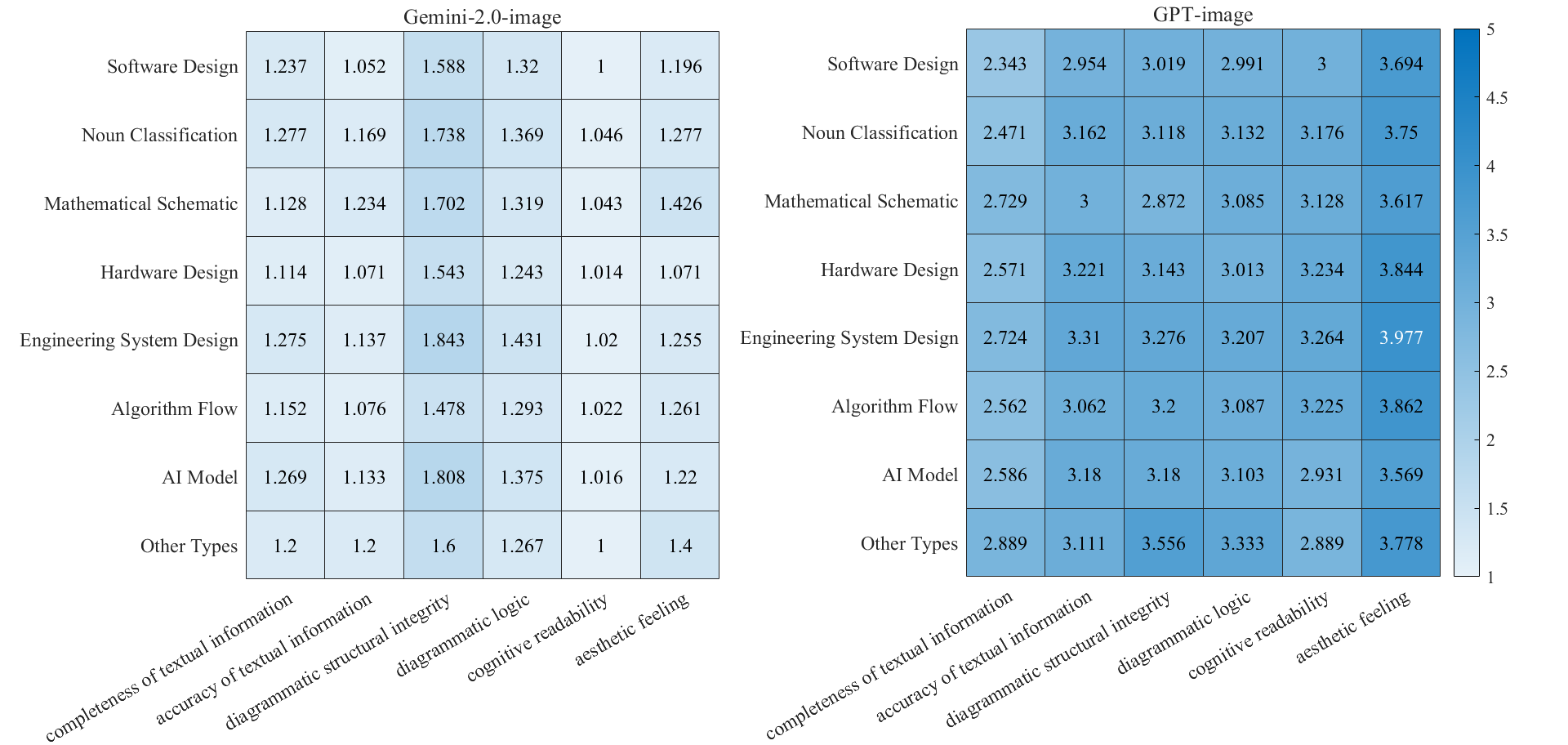} \caption{On different subjects of computer science data, the average score of GPT-4o-image and Gemini-2.0-Flash scores in the six major indicators.
} \label{cs_hm1}   \end{figure}

 As can be seen from Fig.\ref{cs_hm1}, the Gemini-2.0-Flash is still considered to have no preliminary ability to generate scientific maps in computer science, although there is some improvement in the ratings compared to the natural science data. For GPT-4o-image, there was a significant decrease in the scores on the measures of completeness and accuracy of the text information. Compared with the schematics of natural science, the schematics of computer science often have more words and more complex flow structure. This makes the generation of image elements and text elements, GPT-4o-image does not meet the performance of natural science, the same ability to generate images. But at the same time, another noticeable improvement is the ability of GPT-4o-image to be readable and aesthetically pleasing. This is still relevant to the schematic nature of computer science, because most of them are flowcharts, consisting of elements such as text, borders, and arrows. For such diagrams, GPT-4o-image is easier to draw. In contrast to natural science images, generative models need to accurately depict complex and specialized graphical elements such as cellular structures, electron spins, animal organs, and ecosystems. As a result, there is worse performance in natural science diagrams.

\begin{figure}[htbp]  \centering\includegraphics[width=1\textwidth]{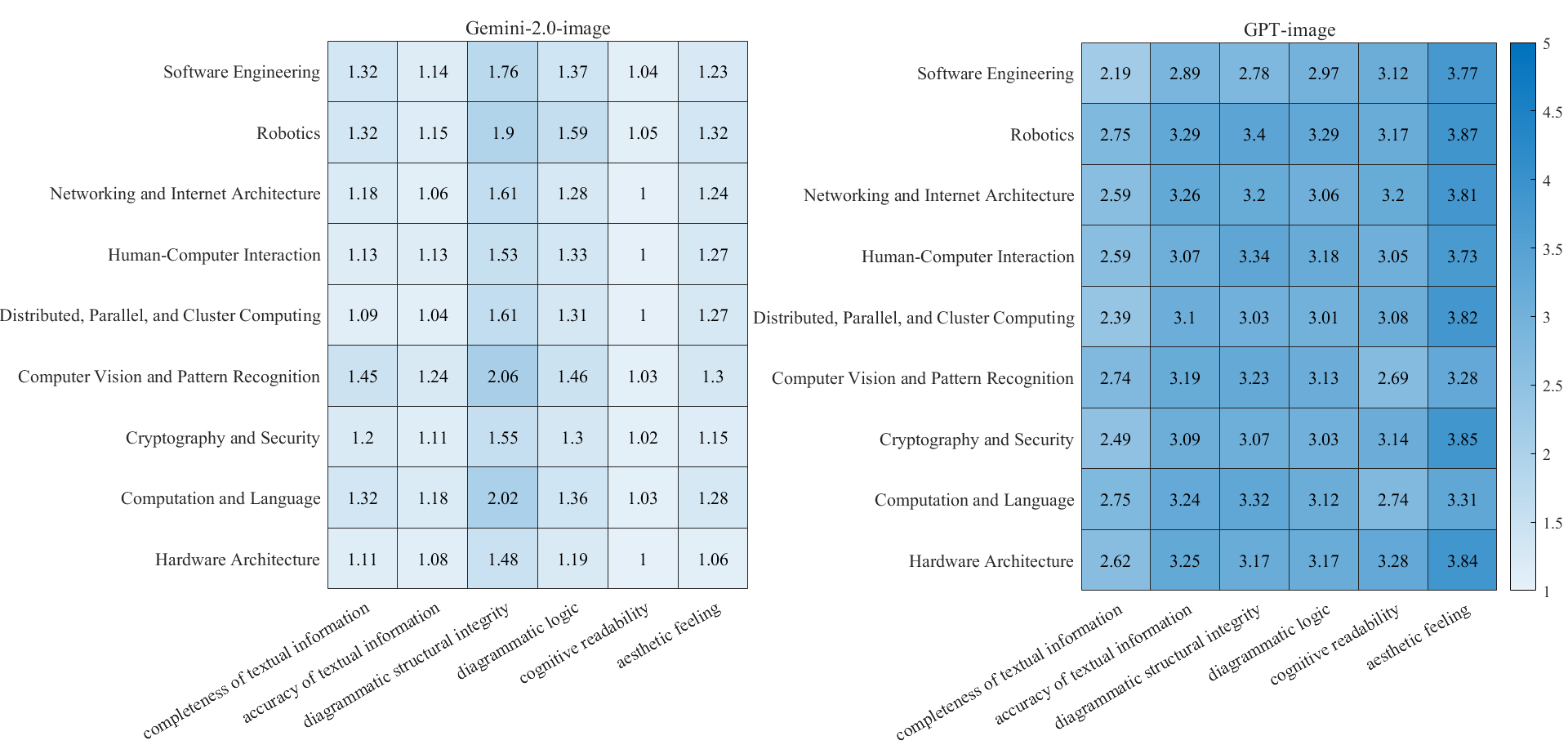} \caption{On different types of computer science data, the average score of GPT-4o-image and Gemini-2.0-Flash scores in the six major indicators.
} \label{cs_hm2}   \end{figure} 

 In different subjects, GPT-4o-image alternative shows absolutely no difference. However,  ``Computer Vision and Pattern Recognition'' and ``Computation and Language'' doesn't score well in terms of brevity and aesthetics. In fact, these two disciplines represent one of the hottest areas of artificial intelligence right now which are computer vision,  pattern recognition and natural language processing. The average level of human drawing in this field is also gradually increasing. Therefore, for the corresponding image generation model, requirement of the quality of the generated image will also be higher.

 \subsection{Analysis of Generation Results}

 \begin{figure}[htbp]  \centering\includegraphics[width=1\textwidth]{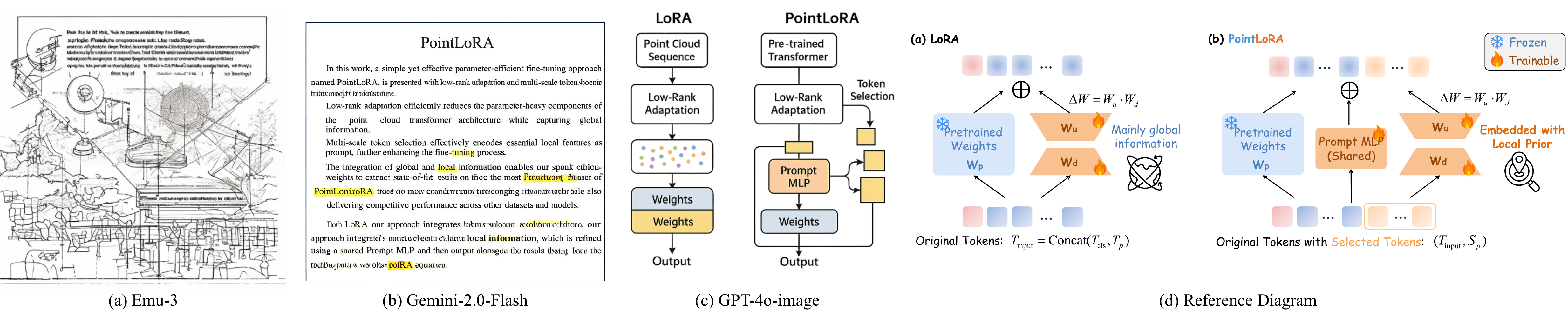} \caption{Computer science paper illustrations generated by different image generation models under the same prompt are compared with the original paper illustration.} \label{vcs}   \end{figure} 

 \begin{figure}[htbp]  \centering\includegraphics[width=1\textwidth]{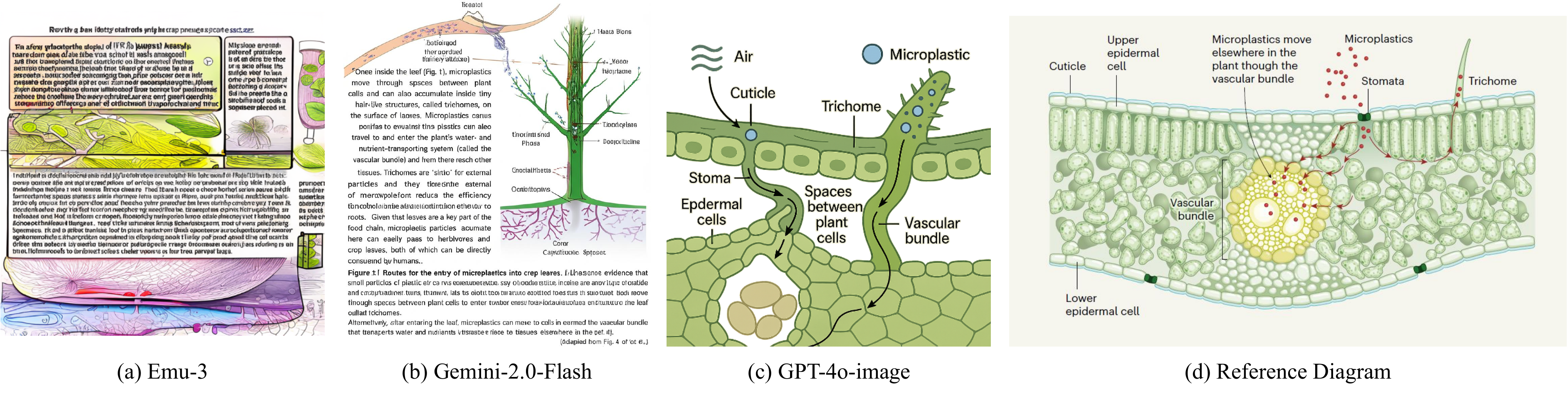} \caption{Natural science paper illustrations generated by different image generation models under the same prompt are compared with the original paper illustration.} \label{vns}   \end{figure} 

Fig.\ref{vcs} and \ref{vns}  show the comparison between the illustrations generated by the three image generation models (Emu-3, Gemini-2.0-Flash, and GPT-4o-image) and the author's original image in the paper. As can be seen, Emu-3 does not have any understanding of scientific writing, and the content it generates is irrelevant to our requirements. Gemini-2.0-Flash simply draws text in an image in Fig.\ref{vcs}. There are no graphic elements, and the text is problematic because they are more like symbols than words. In Fig.\ref{vns},  despite the appearance of the plant-like structure, the resulting image is still difficult to interpret. At the same time, there are a large number of text symbols in the generated illustrations which are similar to the original text of the paper. This situation is also common in other generated illustrations. 

 However, GPT-4o-image has a significant advantage over other models in terms of the quality of content generated. It produces illustrations with well-defined and well-expressed text. The structure of the illustration is clear. The basic elements of the reference image are reflected in the generated results. It can be said that GPT-4o-image has had preliminary, relatively qualified scientific text understanding and image generation capabilities. It can simply and clearly generate scientific, inferential and logical images.  However, this is only preliminary capability. It should be noted that there are still significant problems in generating scientific illustrations from GPT. Examples include missing elements, omissions and errors in textual representation. Compared with the reference images drawn by human experts, there is still a big gap in their correctness and scientific accuracy.

 \begin{figure}[htbp]  \centering\includegraphics[width=1\textwidth]{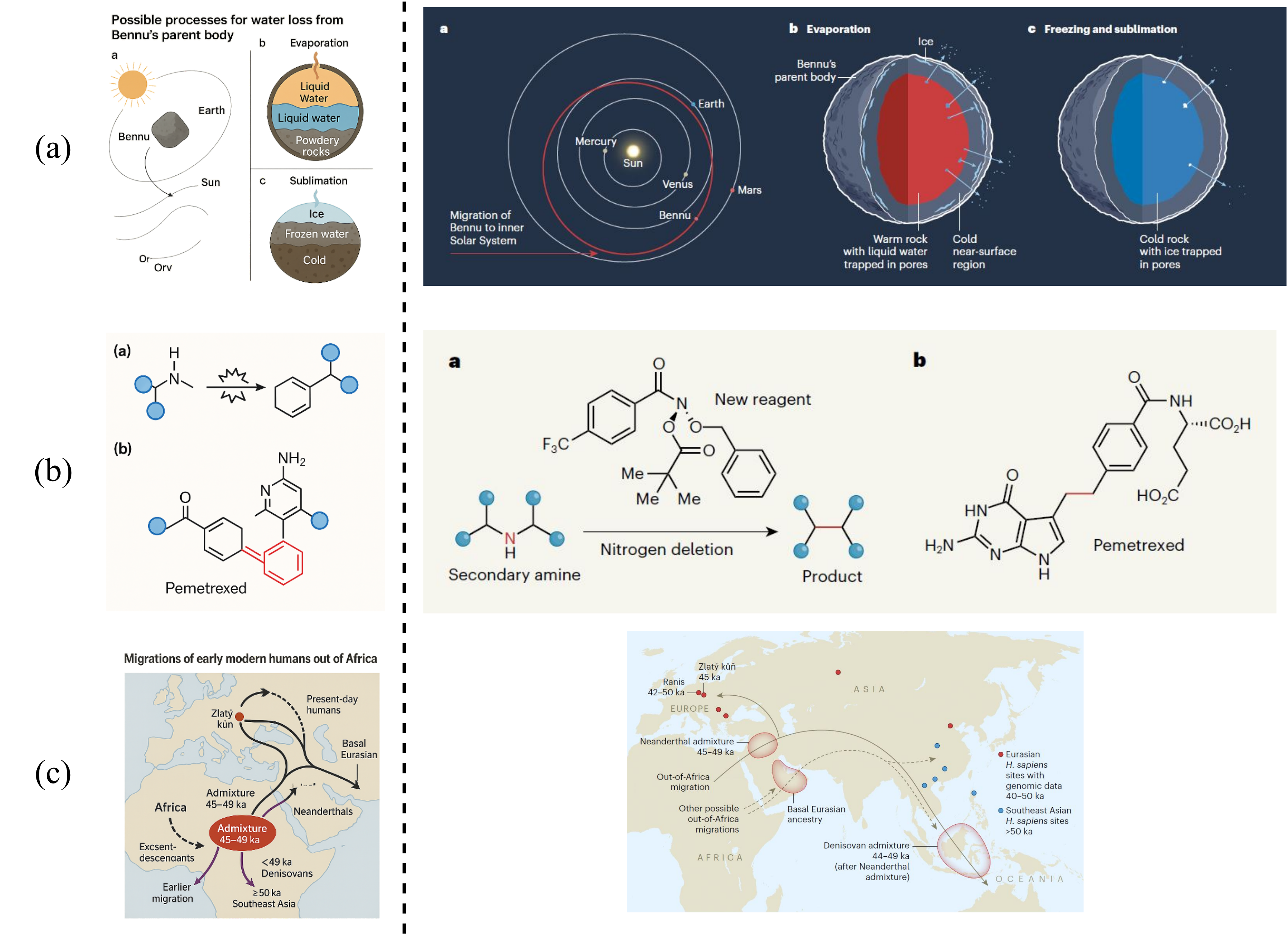} \caption{Illustrations generated by GPT-4o-image (left) and their  reference from original paper (right).} \label{example}   \end{figure} 

 Fig.\ref{example} shows generation results for GPT-4o-image which reflect the common problem of GPT-4o-image. As can be seen from Fig.\ref{example}(a) , the illustrations generated by GPT contain common errors, such as the sun orbiting the Earth. At the same time, although it can draw multi-sub-graphs according to the requirements, the detailed information in the graph is still missing. In Fig.\ref{example}(b) , it is seen that GPT-4o-image has the rudimentary ability to draw structural formulas for organic compounds. But there are still obvious scientific errors in the results, such as the reaction conditions in the diagram. At the same time, the compounds involved in the reactions and the results are not plotted correctly. This shows that the GPT understanding and expression of organic chemistry also has a lot of room for improvement. As you can see from Fig.\ref{example}(c) , GPT-4o-image has rudimentary location understanding and map generation capabilities. However, for more precise positioning and interpretation of geographical processes, GPT-image still has errors and omissions.

\section{Conclusion and Discussion}

In this paper, we propose SridBench, which is the first benchmark of scientific research illustration drawing for image generation model. We propose a strongly inferential drawing scenario called scientific research illustration drawing. Using human experts and MLLM, we meticulously collected and screened triplet data on the scientific paper website for the evaluation of the scientific mapping ability of the image generation model. 1120 triples across 13 disciplines in the natural and computer science were used to test the ability of current scientific research to map multiple graph models. At the same time, we design six indicators to evaluate the generated illustrations. 

We found that, with the exception of GPT-4o-image, other image generation models, such as Gemini-2.0-Flash, do not have any scientific mapping capabilities. GPT-4o-image can preliminarily complete scientific research drawing tasks, generate clear text and complete structure, and have a certain degree of professional results.

However, there are problems with the GPT-4o-image as well. There is a general lack of text information, visual elements are also missing. At the same time, some hallucinations and common sense errors were also found in the GPT-4o-image generation results. This means that, at present, there is still much room for improvement in the ability of scientific research illustration drawing of image generation models. How to improve the generation ability of the image generation model in the task of strong inference should be the focus of the next researchers.

\clearpage
\bibliographystyle{IEEEtran}
\bibliography{main.bib}
\clearpage
\appendix

\section{Prompts used in data process and judgement}\label{prompt}

\textbf{Image Filter} Please strictly judge whether this picture belongs to the concept diagram, model frame diagram, process flow diagram or structure diagram in the academic paper. Return 1 if it is of the following type: concept diagram, model frame diagram, process flow diagram or structure diagram. Return 0 if it is of the following type: experimental result graph, statistical graph, photo, table, mathematical formula, pseudocode. Returns only a single number, 1 or 0, with no explanation.  Note that some of the subplots in the images contain schematics and graphs of statistical analysis and experimental results. In each case, a“0” is returned as long as it contains a statistical analysis of the data and a diagram of the experimental results (not just a diagram).

\textbf{Text Generation(for computer science)} You will get the text of a TEX file. In this file, find the text associated with the image $\{img_name\}$. Note that you can first find the latex figure class where $\{img_name\}$ is located and identify its latex label in label, and then locate the text according to the label. The process cannot be included in the output. The output is the original content associated with $\{img_name\}$. Here is the paper: $\{paper_tex\}$. Output the whole text content of the section (not just the name and label of the section) in which the image is located directly, without anything else.

\textbf{Image Generation} You are a scientist and now you are going to draw a diagram for the computer (natural) science research paper. You will be given the paper section where the diagram is located and the caption of the diagram in the paper. The section is: {section}. The caption is: {caption}. Please draw a professional, rigorous and scientific diagram. You can use different colors and some graphic legends or logos appropriately. Note that the captions we provide do not need to be drawn in the diagram.

\textbf{Image Judgement} You are a researcher who evaluates illustrations in research papers. Next you'll receive two diagrams, the first one by an anthropologist and the second one by an AI model based on the same cue.                      Please rate the graph generated by the AI model on: completeness of textual information (whether it contains all the textual information in the reference graph), accuracy of textual information (whether the textual information is scientifically rigorous), diagrammatic structural integrity (does it draw all the elements of the diagram) , diagrammatic logic (does it arrange the elements scientifically and logically) , cognitive readability (does it allow the reader to understand the content concisely) , aesthetic feeling, ie whether a drawing is aesthetically pleasing or has a sense of design.                     On a scale of 1 to 5(1: fail, 2: poor, 3: fair, 4: good, 5: excellent).                   Please return your comments in the following format: {'completeness of textual information': 4, 'accuracy of textual information': 4, 'diagrammatic structural integrity': 5 , 'diagrammatic logic':4 , 'cognitive readability': 2, 'aesthetic feeling': 3}                    Just return {} and its contents, as in the example above, without returning anything else.                       Here are two pictures: the first is drawn by a human, and the second is drawn by an AI. "


\end{document}